\pdfoutput=1

\documentclass[11pt]{article}

\usepackage{ACL2023}

\usepackage{times}
\usepackage{latexsym}
\usepackage{graphicx}
\usepackage{amsmath}
\usepackage{amssymb}
\usepackage{booktabs}
 \usepackage{mathrsfs}
 \usepackage{bbm}
\usepackage{makecell}
\usepackage{algorithmic}
\usepackage{array}
\usepackage{epsfig}
\usepackage{multirow}
\usepackage{color}
\usepackage{subcaption}
\usepackage{soul}
\usepackage{float}
\makeatletter
\usepackage{hyperref}
\makeatother

\usepackage[T1]{fontenc}

\usepackage[utf8]{inputenc}

\usepackage{microtype}

\usepackage{inconsolata}

\usepackage[normalem]{ulem}
\useunder{\uline}{\ul}{}

\title{Deeply Coupled Cross-Modal Prompt Learning}
\author{
    Xuejing Liu \textsuperscript{\rm1},
    Wei Tang$^+$ \textsuperscript{\rm2},
    Jinghui Lu \textsuperscript{\rm1}, 
    Rui Zhao \textsuperscript{\rm1},
    Zhaojun Guo$^+$ \textsuperscript{\rm3},
    Fei Tan$^*$ \textsuperscript{\rm1}
    \\
    \textsuperscript{1} SenseTime Research \\
    \textsuperscript{2} Nanjing University of Science and Technology \\
    \textsuperscript{3} Fudan University \\
    \texttt{\{liuxuejing, lujinghui1, zhaorui, tanfei\}@sensetime.com} \\
    \texttt{weitang@njust.edu.cn}, \texttt{zhaojun.guo1999@gmail.com}\\
    \texttt{}\\}

\begin{document}
\maketitle
\def\thefootnote{+}\footnotetext{Work was done during internship at SenseTime Research}
\def\thefootnote{*}\footnotetext{Corresponding author}
\begin{abstract}
Recent advancements in multimodal foundation models (e.g., CLIP) have excelled in zero-shot generalization. Prompt tuning involved in the knowledge transfer from foundation models to downstream tasks has gained significant attention recently. Existing prompt-tuning methods in cross-modal learning, however, either solely focus on language branch, or learn vision-language interaction in a shallow mechanism. In this context, we propose a \textbf{D}eeply coupled \textbf{C}ross-modal \textbf{P}rompt learning (DCP) method based on CLIP. DCP flexibly accommodates the interplay between vision and language with a \textbf{C}ross-\textbf{M}odal \textbf{P}rompt \textbf{A}ttention (CMPA) mechanism, which enables the mutual exchange of respective representation through a well-connected multi-head attention module progressively and strongly. We then conduct comprehensive few-shot learning experiments on 11 image classification datasets and analyze the robustness to domain shift as well. Thorough experimental analysis evidently demonstrates the superb few-shot generalization and compelling domain adaption capacity of a well-executed DCP. The code can be found at \href{https://github.com/GingL/CMPA}{https://github.com/GingL/CMPA}.

\end{abstract}

\section{Introduction}
Large foundation models pre-trained on web-scale image-text pairs such as CLIP~\cite{DBLP:conf/icml/RadfordKHRGASAM21} and ALIGN~\cite{DBLP:conf/icml/JiaYXCPPLSLD21} have shown promising performance on zero-shot image classification. Research has repeatedly shown that the general knowledge learned by the foundation models can also be transferred to diverse downstream tasks, such as few-shot image classification~\cite{DBLP:journals/ijcv/ZhouYLL22,DBLP:conf/cvpr/ZhouYL022}, visual grounding~\cite{DBLP:conf/acl/SubramanianMD0022}, visual question answering~\cite{DBLP:conf/ijcai/Liu0P022} and so on. They have exhibited a significant potential in open-vocabulary scenarios. Thus, the challenge associated with how to efficiently and effectively adapt large pre-trained models to downstream tasks has garnered increasing attention especially in low-resource training scenarios.

Directly fine-tuning the foundation model is infeasible due to the massive training parameters and the catastrophic forgetting caused by overfitting~\cite{DBLP:journals/corr/KirkpatrickPRVD16}. 
In contrast, the parameter-efficient \emph{prompt tuning} approach explored in natural language processing has yielded significant success~\cite{DBLP:conf/emnlp/LesterAC21}, leading to an increased examination of this technique within the realm of multi-modality, especially in the language-branch of CLIP.
For example,
CoOp~\cite{DBLP:journals/ijcv/ZhouYLL22} and ProDA~\cite{DBLP:conf/cvpr/LuLZL022} explore the vanilla few-shot learning based on CLIP by adjusting the embedding or distribution of the text prompt. 
CoCoOp~\cite{DBLP:conf/cvpr/ZhouYL022} and ProGrad~\cite{DBLP:journals/corr/abs-2205-14865} focus more on the unseen classes. They contextualize the text prompt either under the supervision of visual clues or tweak gradient direction to improve the generalization ability of the model.

The aforementioned approaches, however,
only adjust the text embedding of CLIP and neglect the visual branch. The success of VPT~\cite{DBLP:conf/eccv/JiaTCCBHL22} demonstrates the effectiveness of visual prompt learning. Inspired by this work, UPT~\cite{DBLP:journals/corr/abs-2210-07225} and MaPLe~\cite{DBLP:journals/corr/abs-2210-03117} synergize the visual and textual prompts. 
Specifically, UPT improves the few-shot learning ability by generating visual and text prompts initially. MaPLe achieves better performance in the classification of unseen classes. 
They uncover the underlying rationale and limitations of dual-branch prompt tuning.

Concretely, the dual-branch CLIP learns the visual and language synergy only based on contrastive learning, whereas both branches lack mutual communication at the early stage of the network. Multi-modal prompt learning techniques, such as MaPLe and UPT, incorporate language-vision interactions of the network and achieve substantially improved performance, highlighting the significance of the cross-modal interactions. However, previous studies have leveraged language-vision interactions at a superficial level. For example, UPT generates visual and text prompts before they are fed into the corresponding encoders. MaPLe generates visual prompts conditioned on language counterparts by a mapping function. 
Many studies~\cite{DBLP:conf/iclr/DosovitskiyB0WZ21,DBLP:conf/icml/WangYMLBLMZZY22} have shown that neural networks, especially transformer-based models, can leverage the deep fusion of information from multiple views to improve their performance. It remains less explored in the thread of multi-modal few-shot learning. 
To this end, we design \textbf{D}eeply coupled \textbf{C}ross-modal \textbf{P}rompt learning (DCP) enhancing the language-vision interaction. Specifically, DCP is built upon CLIP, with additional text and visual prompts across multiple layers. 
Different from previous methods with deep prompt tuning~\cite{DBLP:conf/eccv/JiaTCCBHL22,DBLP:journals/corr/abs-2210-07225,DBLP:journals/corr/abs-2210-03117}, DCP only initializes the first layer of visual and text prompt randomly. The subsequent prompts are generated by \textbf{C}ross-\textbf{M}odal \textbf{P}rompt \textbf{A}ttention (CMPA) module, which elegantly integrates the prompts from the preceding cross-modal layer. CMPA is characterized with stronger connection in two folds, i.e., \emph{Depth} and \emph{Breadth}. 1) \emph{Depth} means that CMPA intensifies the correlation of the prompts among different layers. 2) \emph{Breadth} refers to that CMPA amplifies the interaction between visual and language modalities. CMPA is the core module to realize the deep coupling between two modalities.
Essentially, DCP empowered by CMPA amalgamates uni-branch and dual-branch multi-modal pre-training paradigms in a favorable way in an attempt to bridge the discrepancy between visual and textual knowledge without introducing too much overhead.

To conclude, the contributions of this work are as follows:
\begin{itemize}
    \item We develop a deeply coupled cross-modal prompt learning (DCP) method with a core module cross-modal prompt attention (CMPA). CMPA can reinforce the interaction between visual and language modals across different layers. 
    \item We benchmark our method on 11 image classification datasets consisting of generic objects, scenes, actions and fine-grained categories. Our method surpasses visual prompt tuning, text prompt tuning and existing competitive multi-modal prompt tuning methods under the few-shot setting.
    \item We conduct experiments on domain adaptation tasks. Our method achieves comparable performance to the state-of-the-art methods, indicating the robustness of our method to domain shift. 
\end{itemize}

\section{Related Work}

\subsection{Vision-language Pre-trained Models}
The advent of Transformer~\cite{DBLP:conf/nips/VaswaniSPUJGKP17} has accelerated the development of large-scale pre-training.
The application of Transformer in the multi-modal is divided into two schools of thought: one is the single-stream model, in which language and vision information are fused at the beginning and fed directly into the encoder together; the other is the dual-stream model, in which language and vision information first pass through two separate encoder modules at the beginning, and then the different modal information is fused through the cross Transformer.

At the outset, the basic architecture of some contemporaneous work is BERT. The images are detected with Faster-RCNN~\cite{DBLP:conf/nips/RenHGS15} for region features, and these image region features are fed into BERT along with text information to align the text and image information. Following the same process as BERT, these methods first pre-train and then fine-tune on the corresponding tasks. 
Single-stream networks~\cite{DBLP:journals/corr/abs-1908-03557,DBLP:conf/emnlp/AlbertiLCR19,DBLP:journals/corr/abs-1909-11740,DBLP:conf/aaai/LiDFGJ20,DBLP:conf/iclr/SuZCLLWD20,DBLP:conf/aaai/ZhouPZHCG20,DBLP:journals/corr/abs-2001-07966,DBLP:conf/cvpr/LuGRPL20} fuse information from different modalities directly through an encoder.
The dual-stream models~\cite{DBLP:conf/nips/LuBPL19,DBLP:conf/emnlp/TanB19} integrate different modal information through cross modal transformer. 
Empirically single-stream networks are more sufficient for information fusion, while dual-stream networks can be more efficient for training due to fewer training parameters. 
In the design of our method, we aim to combine the advantages of the single-stream and dual-stream, so as to enhance the cross-modal integration without introducing many training parameters.

Recent cross-modal large-scale pre-training models have made greater breakthroughs in training data scale and tasks by devising various model architectures and training objectives, and have achieved impressive performance in many downstream tasks. CLIP~\cite{DBLP:conf/icml/RadfordKHRGASAM21} and ALIGN~\cite{DBLP:conf/icml/JiaYXCPPLSLD21} got remarkable zero-shot results after being pre-trained on millions or billions of (image, text) pairs collected from the internet. Coca~\cite{DBLP:journals/corr/abs-2205-01917} combined the advantages of the contrast learning method~\cite{DBLP:conf/icml/RadfordKHRGASAM21} and the generative model SiMVLM~\cite{DBLP:conf/iclr/WangYYDT022} by adding caption loss to the contrast loss of CLIP.
OFA~\cite{DBLP:conf/icml/WangYMLBLMZZY22}, Unified-IO~\cite{DBLP:journals/corr/abs-2206-08916} and Florence~\cite{DBLP:journals/corr/abs-2111-11432} unified vision, language and multi-modal tasked by pre-training on both cross-modal and uni-modal data. These methods have achieved state-of-the-art results in many downstream tasks.
Some methods are dedicated to improving the performance of certain specific tasks.
UniTAB~\cite{DBLP:conf/eccv/YangGW000LW22} focused on grounded vision-language tasks such as grounded captioning and visual grounding. GLIP~\cite{DBLP:conf/cvpr/LiZZYLZWYZHCG22} unified object detection and phrase grounding for pre-training.
Pre-training models have opened up a situation where deep learning models scale and perform in tandem, becoming a revolutionary breakthrough in artificial intelligence and deep learning.

\subsection{Prompt Learning}
For a long time, first pre-training then fine-tuning was the dominant approach to apply large foundation models to downstream tasks. However, fine-tuning for large models is inefficient and may cause catastrophic forgetting~\cite{DBLP:journals/corr/KirkpatrickPRVD16}. Prompt learning is proposed to address the above problems. The prompt is usually a series of trainable parameters inserted into the input. 
The success of prompt learning in NLP~\cite{DBLP:conf/emnlp/LesterAC21} has inspired its application in other modalities. 
VPT~\cite{DBLP:conf/eccv/JiaTCCBHL22} is a typical successful application of prompt learning on computer vision. Prompt learning has generated more attention and made great progress in cross-modal learning.

SoftCPT~\cite{DBLP:journals/corr/abs-2208-13474} and CPL~\cite{DBLP:journals/corr/abs-2210-10362} applied prompt tuning to different vision and language tasks and outperformed single-task prompt tuning method.
CoOp~\cite{DBLP:journals/ijcv/ZhouYLL22}, ProDA~\cite{DBLP:conf/cvpr/LuLZL022} and UPT~\cite{DBLP:journals/corr/abs-2210-07225} adapted prompt learning to traditional few-shot visual recognition with CLIP as the backbone. CoCoOp~\cite{DBLP:conf/cvpr/ZhouYL022}, ProGrad~\cite{DBLP:journals/corr/abs-2205-14865} and MaPLe~\cite{DBLP:journals/corr/abs-2210-03117} improved the classification performance of pre-trained models on novel categories by prompt learning. 
Different from previous methods, our approach brings stronger connection between modalities and layers with proposed cross-modal prompt attention. The stronger interaction between vision and language enables our method to get state-of-the-art performance in the few-shot learning.

\section{Method}
In this section, we first introduce the preliminaries, including CLIP~\cite{DBLP:conf/icml/RadfordKHRGASAM21}, CoOp~\cite{DBLP:journals/ijcv/ZhouYLL22} and VPT~\cite{DBLP:conf/eccv/JiaTCCBHL22}. Then, we describe our deeply coupled prompt learning (DCP) and detail its underlying module CMPA.

\begin{figure*}[tp]
	\centering
	\includegraphics[width=0.95\textwidth]{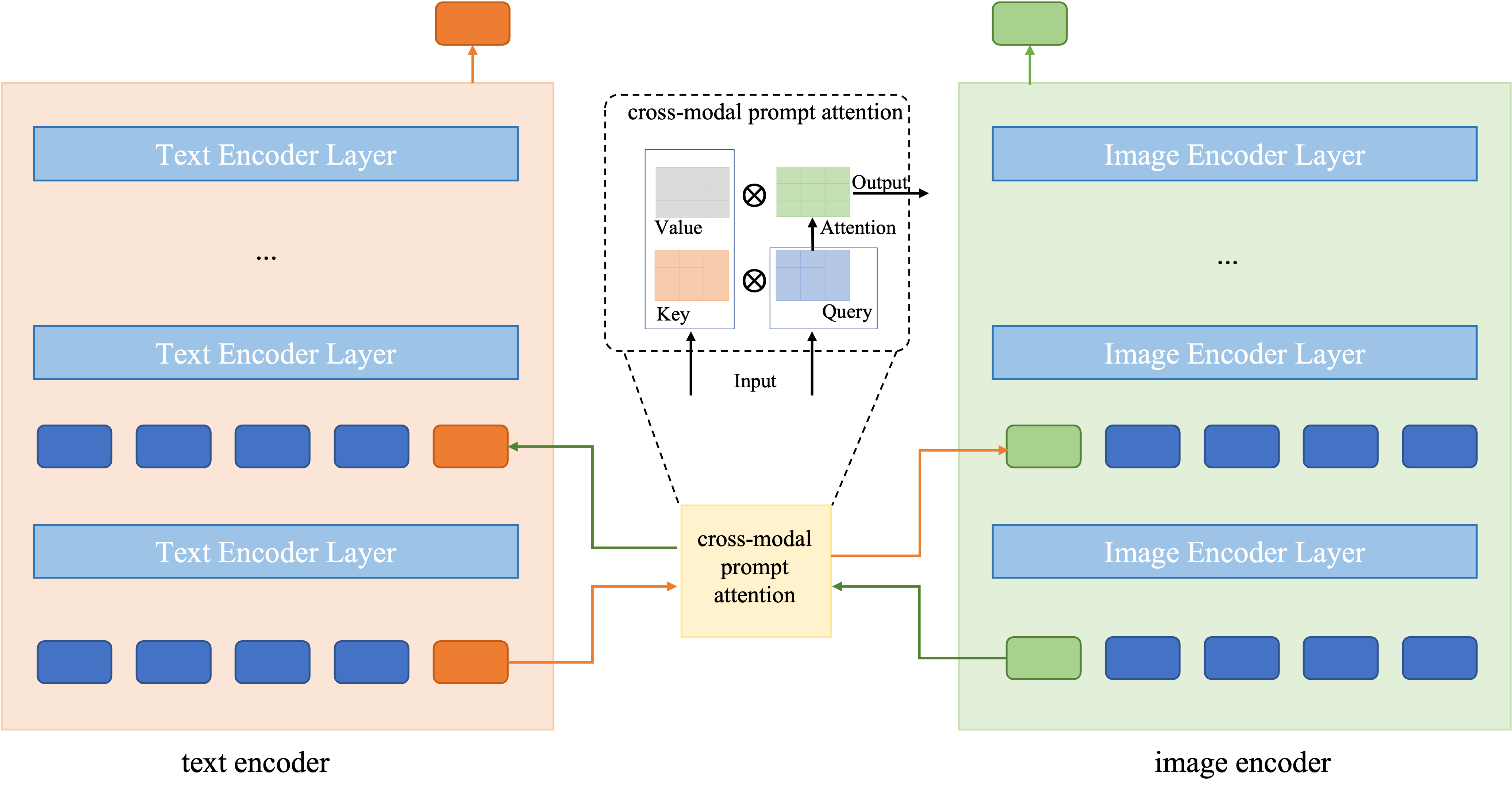}
	\caption{The architecture of deeply coupled prompt learning and cross-modal prompt attention module.}
	\label{fig:pipeline}
\end{figure*}

\subsection{Preliminaries}
\paragraph{CLIP}
is a dual-encoder pre-trained model which consists of a text encoder and an image encoder. The text and image are independently encoded by the corresponding encoder, then projected to the same embedding space by a projection layer. Specifically, the backbone of the image encoder is ResNet~\cite{DBLP:conf/cvpr/HeZRS16} (d=256) or ViT (d=512), which can map the high-dimension image into a low-dimension embedding.  The text encoder is built based on the decoder of Transformer~\cite{DBLP:conf/nips/VaswaniSPUJGKP17}, which is also known as GPT~\cite{DBLP:conf/nips/BrownMRSKDNSSAA20}, to generate a vectorized representation for a sequence of words. 
The model uses a contrastive loss to align the two modalities during training stage.  
The training objective is to maximize the cosine similarity for the match image-text pairs and minimize the unmatched ones.

In zero-shot image recognition, the image encoder of CLIP encodes the image into a feature representation $\boldsymbol{x}$. The input text is usually in the form of ``a photo of a \{class\}.'' (\emph{discrete prompt}), where the ``\{class\}'' token is the name of each category. For each dataset containing $K$ categories, a set of text prompts $\{\boldsymbol{w_i}\}_{i=1}^K$ are generated by the text encoder. The prediction probability is computed as
\begin{equation}
p(y \mid \boldsymbol{x})=\frac{\exp \left(\operatorname{cos}\left(\boldsymbol{x}, \boldsymbol{w}_y\right) / \tau\right)}{\sum_{i=1}^K \exp \left(\operatorname{cos}\left(\boldsymbol{x}, \boldsymbol{w}_i\right) / \tau\right)},
\end{equation}
where $\tau$ is a temperature parameter.

\paragraph{CoOp} adapts CLIP to downstream tasks with prompt tuning. Specifically, CoOp tries to learn prompt embedding \emph{(continuous prompt)} during few-shot training to avoid manual prompts. The prompt fed in the text encoder is designed as $t=[V]_1[V]_2...[V]_M[CLASS]$, where $[V]_m \ (m \in \{1,...,M\})$ is initialized with the same dimension as word embeddings. The parameters of the CLIP model is frozen while the prompt is trainable. The prediction probability of CoOp is
\begin{equation}
p(y \mid \boldsymbol{x})=\frac{\exp \left(\operatorname{cos}\left(\boldsymbol{x}, g(\boldsymbol{t}_y\right)) / \tau\right)}{\sum_{i=1}^K \exp \left(\operatorname{cos}\left(\boldsymbol{x}, g(\boldsymbol{t}_i\right)) / \tau\right)},
\end{equation}
where $g(\cdot)$ denotes the text encoder.

\paragraph{VPT} is an efficient and effective way to adapt large-scale Transformer models in vision with only a small amount of trainable parameters. The backbone of VPT is ViT, which is the same as the image encoder of CLIP. There are two variants of VPT: VPT-Shallow and VPT-Deep. VPT-Shallow only inserts prompts into the first layer of the Transformer. The visual prompt can be defined as $p=[P]_1[P]_2...[P]_N$, where $[P]_n \ (n \in \{1,...,N\})$ keeps the same dimension as the image embedding. The input of VPT-shallow is $[x_{cls}, p, x]$, where $x_{cls}$ is the classification token $[CLS]$. VPT-Deep introduces visual prompts at every Transformer layer. The deep VPT can be formulated as
\begin{equation}
\begin{aligned}
{\left[\mathbf{x}_{cls}^i, \ldots, \mathbf{x}^i\right] } & =L^i\left(\left[\mathbf{x}_{cls}^{i-1}, \mathbf{p}^{i-1}, \mathbf{x}^{i-1}\right]\right) \\
i &=1,2, ..., L \\
\mathbf{y} & =\operatorname{Head}\left(\mathbf{x}_{cls}^L\right),
\end{aligned}
\end{equation}
where $L$ denotes the number of Transformer layers and $Head$ is the classification head. Only the prompts and classification head is learnt during training. VPT achieves impressive performance on 24 downstream recognition tasks.


\subsection{Cross-modal Prompt Attention}
Inspired by the advance of prompt learning in vision and language, 
recent studies start to explore multi-modal prompt learning~\cite{DBLP:journals/corr/abs-2210-07225,DBLP:journals/corr/abs-2210-03117}. These methods update the visual and text prompt simultaneously to achieve balance in the learning of visual and text embedding. Although the visual and text embedding are adapted to the few-shot data, the interaction between visual and text is still insufficient. Hence we propose deeply coupled cross-modal prompt learning (DCP), which can enhance the communication between prompts across different layers and modalities. The essential module of DCP is cross-modal prompt attention, which fuses visual and text with multi-head cross-modal attention. Figure~\ref{fig:pipeline} depicts the pipeline of DCP and the detailed architecture of cross-modal prompt attention (CMPA).

Our method follows the implementation of CLIP, which is also a dual-encoder model. Differently, we add prompts to every branch, and enable information fusion between vision and language during training through CMPA. Specifically, CMPA is a multi-head attention with visual and text prompts as inputs. 
The language prompts of the first layer are initialized with the pre-trained CLIP word embeddings of the template 'a photo of a <class>', whereas the visual prompts inserted into the first layer are randomly initialized from a normal distribution. 
Then, the prompts of the next layer are generated by CMPA based on the prompts from the preceding layer. Formally, CMPA can be formulated as 

\begin{align}
\text P_t^{l+1}&=\operatorname{softmax}\left(\frac{P_v^l (P_t^l)^T}{\sqrt{d_k}}\right) P_t^l \\
\text P_v^{l+1}&=\operatorname{softmax}\left(\frac{P_t^l (P_v^l)^T}{\sqrt{d_k}}\right) Pv^l \\
l&=1, 2, ..., N-1 ,
\end{align} 
where $P_t^l$ and $P_v^l$ denote the text prompt and visual prompt the the $l$ layer of each encoder, respectively. $N$ is the depth of CMPA, which is smaller than the length of text and visual encoder. $d_k$ is the dimension of keys.

Different from previous methods, only the prompts from the first layer are randomly generated. The subsequent prompts condition on the prompts from both visual and language modal. CMPA enables information communication between vision and text through corresponding prompts. Totally, CMPA brings stronger feature fusion from two aspects: layers and modalities. Note that CMPA shares parameters from different layers, and the additional trainable parameters is only in a small amount.

\section{Experiments}

\begin{figure*}[t!]
\centering
\includegraphics[width=\textwidth]{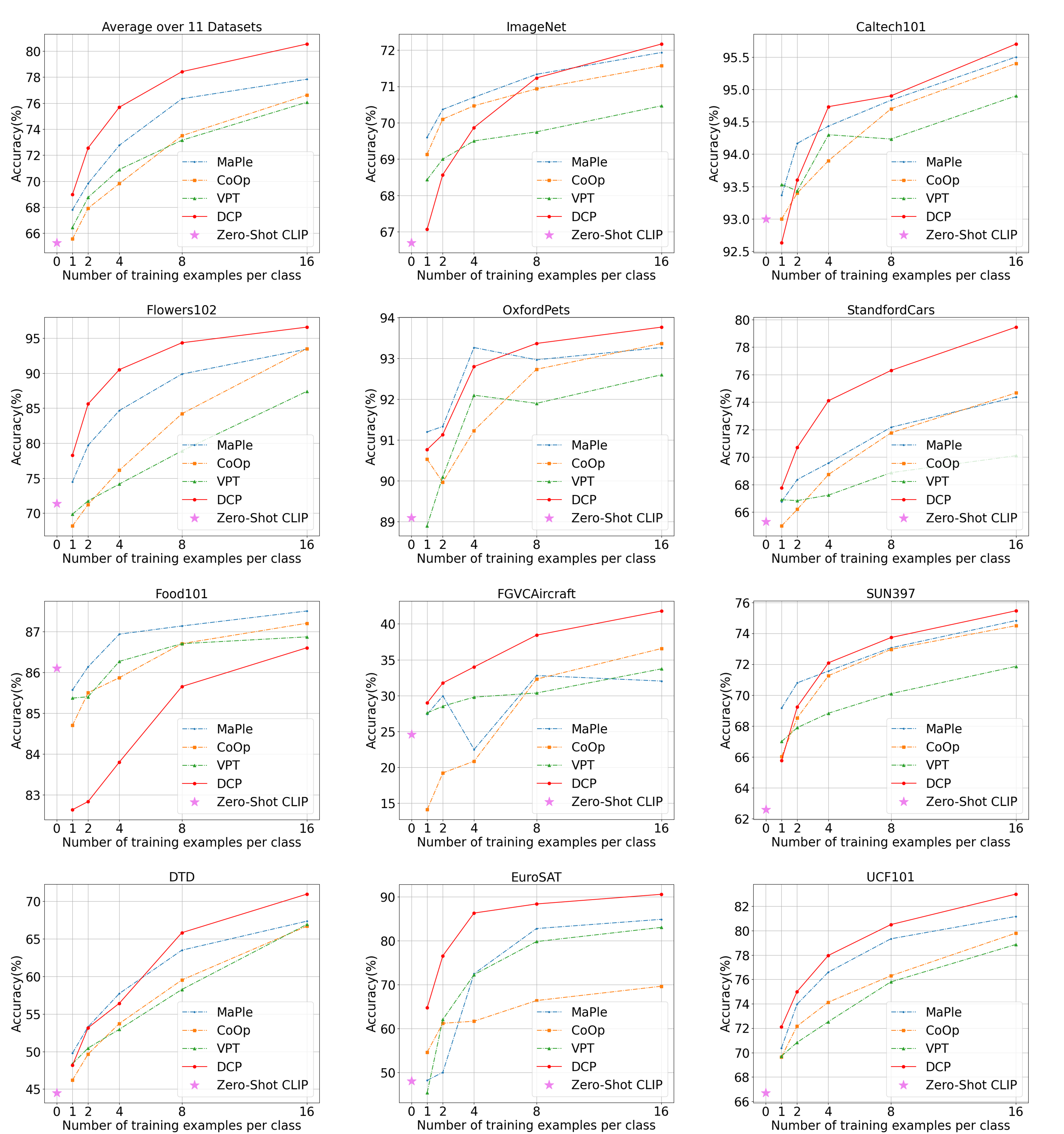}
\caption{\textbf{Main results of few-shot image classification on 11 datasets.} The accuracy (\%) is the average over three runs on 1/2/4/8/16 shots. Overall, our DCP ({\color{red} red line}) outperforms other methods by a large margin on the average results of 11 datasets.}
\label{fig:fewshot_results}
\end{figure*}

In this section, we conduct experiments to evaluate the effectiveness of our method under two settings. One is few-shot visual recognition including 11 different datasets covering generic objects, scenes, actions and fine-grained categories. The other is domain adaptation, where we train our model on ImageNet and evaluate it on other four datasets.

\subsection{Few-shot Learning}
\subsubsection{Datasets}
Following CoOp~\cite{DBLP:conf/emnlp/LesterAC21}, we evaluate our method on 11 public visual recognition datasets: ImageNet~\cite{DBLP:conf/cvpr/DengDSLL009}, Caltech101~\cite{DBLP:conf/cvpr/LiFP04}, OxfordPets~\cite{DBLP:conf/cvpr/ParkhiVZJ12}, StanfordCars~\cite{DBLP:conf/iccvw/Krause0DF13}, Flowers102~\cite{DBLP:conf/icvgip/NilsbackZ08}, Food101~\cite{DBLP:conf/eccv/BossardGG14}, FGVCAircraft~\cite{DBLP:journals/corr/MajiRKBV13}, SUN397~\cite{DBLP:conf/cvpr/XiaoHEOT10}, DTD~\cite{DBLP:conf/cvpr/CimpoiMKMV14}, EuroSAT~\cite{DBLP:journals/staeors/HelberBDB19} and UCF101~\cite{DBLP:journals/corr/abs-1212-0402}. 
We also use the same 1, 2, 4, 8 and 16 shots as CoOp for training and the full test set for evaluation purpose. The reported results are the average over three runs with different random seeds.

\subsubsection{Implementation Details}
We use the pre-trained ViT-B/16 CLIP model as our backbone. The length of prompt tokens for visual and textual context are both 16. The prompt depth is 9 as a trade-off between accuracy and training efficiency. We set the batch-size to 4 with a learning rate of 0.0035 via SGD optimizer. We use 20 epochs for most datasets, except ImageNet, SUN397 and Food101. 
Also, 5-epoch setting works for diverse shots of Food101, 1/2/4-shot of ImageNet, and 1/2-shot of SUN397, respectively.

\subsubsection{Main Results}
  



\paragraph{Baseline Methods.}
We compare our method with the original zero-shot CLIP, text prompt learning (CoOp), visual prompt learning (VPT) and multi-modal prompt learning (MaPLe), which all have ViT-B/16 as visual backbone. Basically, we follow the implementation of MaPLe~\cite{DBLP:journals/corr/abs-2210-03117}. The prompt length of CoOp is set to 16. 
VPT uses a prompt length of 8 and the visual and text prompt length of MaPLe is 2. The training epoch of CoOp is defined as 10, and that of VPT and MaPLe is 5. 
We use the deep variant of VPT in few-shot experiments.
The prompt depth of MaPLe is 9 as their original setting.

\paragraph{Performance Analysis.}
Figure~\ref{fig:fewshot_results} demonstrates our results comparison with other methods. The top left sub-figure shows the average performance of four methods. We can have the following findings. 
1) Overall, cross-modal prompt learning (DCP and MaPLe) gets a large performance gain compared with single-modal prompt learning methods (VPT and CoOp). VPT and CoOp achieve comparable performance on different shots. These results demonstrate the superiority of cross-modal prompt learning over uni-modal prompt learning.
2) Although both belong to multi-modal prompt learning methods, our method still outperforms MaPLe on 1/2/4/8/16 shots settings by 1.72/3.18/3.19/2.20/2.76(\%). MaPLe utilized a linear layer to generate visual prompts from text prompts. Our proposed DCP enhances the interaction between vision and language with a cross-modal prompt attention, which can not only guide visual embedding learning through text prompts, but also influence the language embedding with visual prompts. 
3) Compared with 2/4/8/16 shots, our approach achieves a lower performance gain on one shot. We can also find that on separate datasets, our method achieves the best performance in almost all 16-shot cases (except for Food101).
This phenomenon indicates that our method is more effective in cases where the number of shots is relatively large. This is probably because the alignment between different modals is more challenging due to the small number of samples per category.

For individual datasets, we find that our approach has significant performance improvements on Flowers102, StanfordCars, FGVCAircraft, and EuroSAT. However, on 
the datasets of general categories such as ImageNet and Caltech101, our method does not achieve satisfactory performance when the number of shots is less than 16. We can conclude that our method is more robust for fine-grained classification datasets, and we need more shots for general category classification.
On the dataset of Food101, our method performs slightly lower than MaPLe. We also find that all methods underperform zero-shot on 1-shot setting. 
We suppose this phenomenon comes from the noisy training data of Food101~\cite{DBLP:conf/eccv/BossardGG14}.

\subsubsection{Ablation Study}
The are two important settings in CMPA: the feature fusion method in different prompts and parameter sharing of CMPA across different layers. We conduct corresponding ablation experiments in this section to find the optimal setting.
\paragraph{Feature Fusion in Prompts.}
\begin{figure}[tp]
	\centering
	\includegraphics[width=0.45\textwidth]{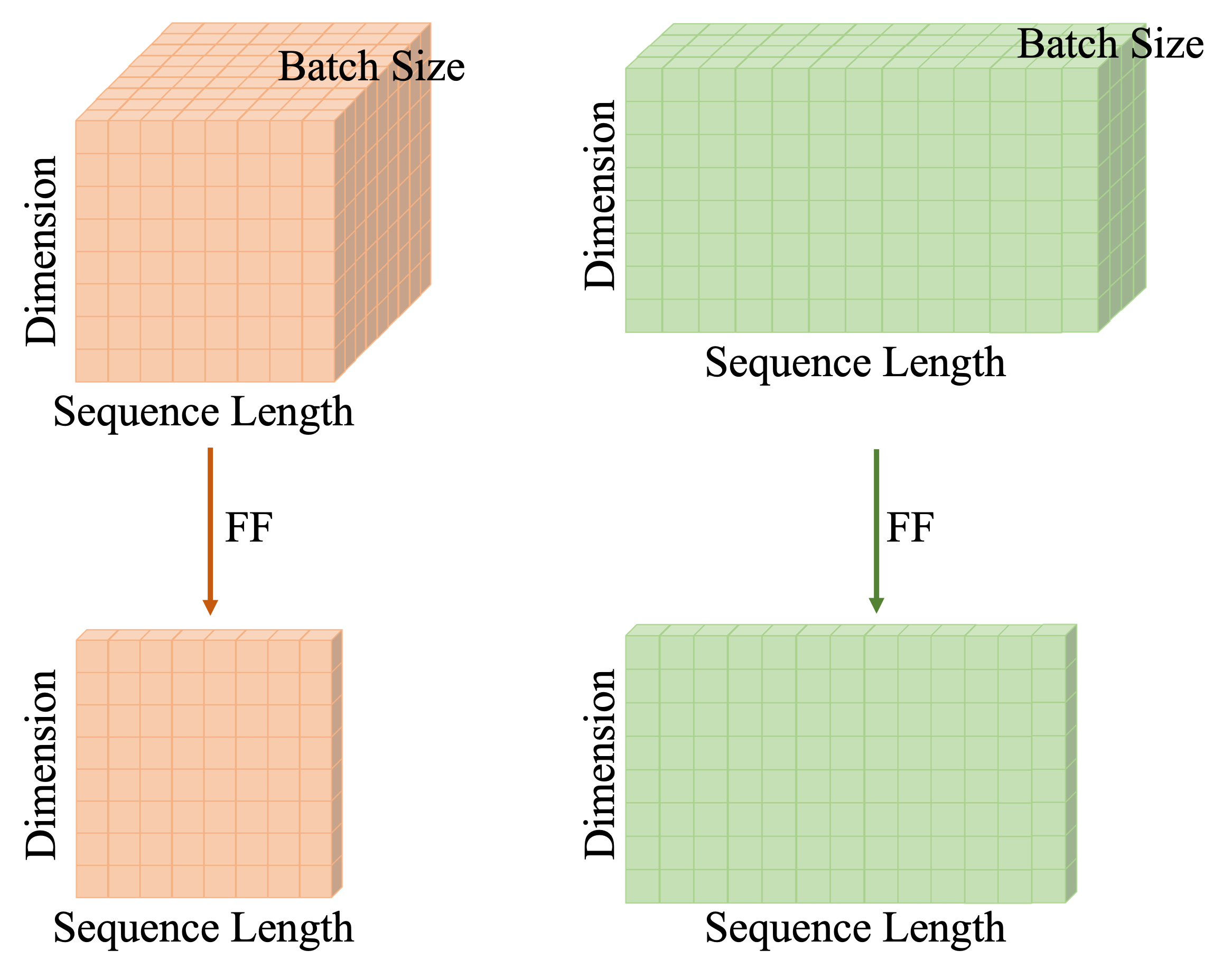}
	\caption{The illustration of feature fusion (FF). The left branch represents the text prompt, and the right shows the visual prompt.}	\label{fig:fuse_sketch}
\end{figure}
\begin{figure}[tp]
	\centering
	\includegraphics[width=0.48\textwidth]{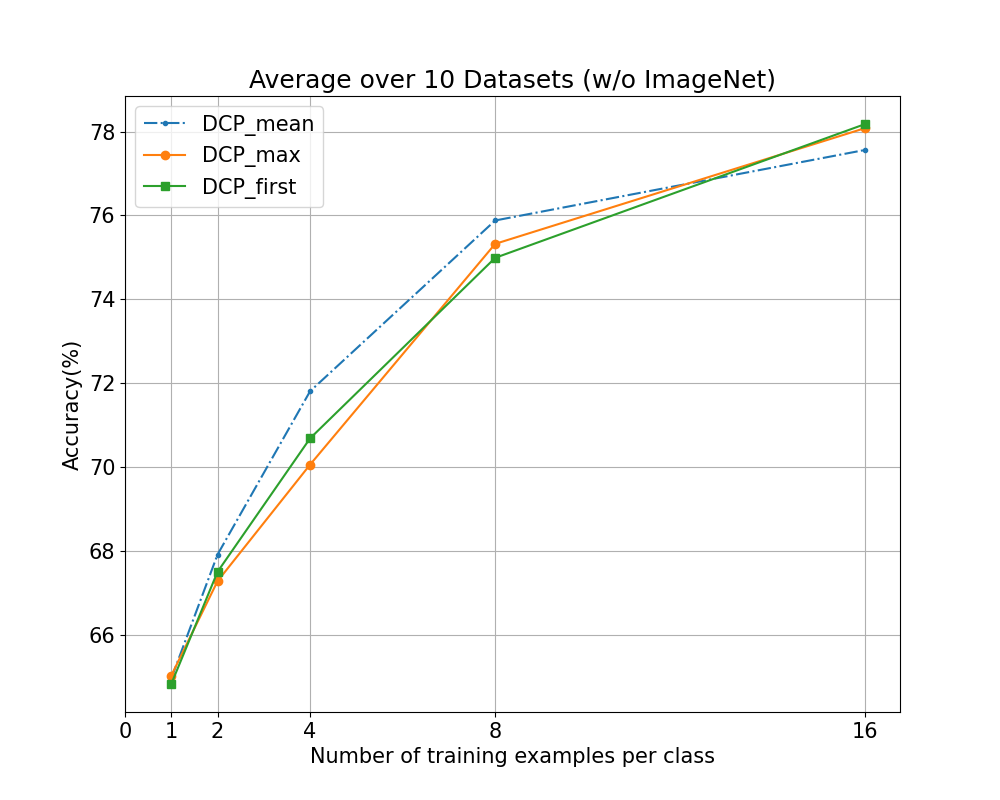}
	\caption{The comparison of different feature fusion methods on 10 datasets without ImageNet.}
	\label{fig:feat_fuse}
\end{figure}
Before the visual and text prompts are fed into the CMPA, the dimension of the batch size is supposed to be consistent. The defined batch size only affects visual prompt while the batch size of text prompts is actually the number of the dataset due to the implementation of CLIP. The dimension transformation of visual and text prompts is shown in Figure~\ref{fig:fuse_sketch}. The batch size of text prompt is actually the number of categories in the dataset.
We experiment with three settings to align the batch size of visual and text prompts. 
Figure~\ref{fig:feat_fuse} reports the average accuracy over three runs on different shots (1/2/4/8/16) of 10 datasets (without ImageNet for time efficiency).
`Avg' means that we use the average of visual and text prompts across the dimension of batch.
`Max' stands for using the features with the highest response across the batch dimension as the visual and text prompt. 
`First' represents that we select the first embedding across the batch dimension of visual and text prompts to feed into CMPA.
Overall, the `avg' setting of feature fusion can achieve better performance compared with `max' and `first'.
\paragraph{Parameter Sharing.}
\begin{table}[]
\begin{tabular}{lccccc}
\toprule
\textbf{Variant} & \textbf{2}& \textbf{4}& \textbf{6}& \textbf{8}& \textbf{16}      \\ \midrule
w/ PS& 68.99   & 72.56   & 75.69   & 78.42   & 80.55   \\
w/o PS& 67.42   & 71.34   & 75.27   & 78.49   & 80.53   \\ \bottomrule
\multicolumn{1}{l}{} & \multicolumn{1}{l}{} & \multicolumn{1}{l}{} & \multicolumn{1}{l}{} & \multicolumn{1}{l}{} & \multicolumn{1}{l}{}
\end{tabular}
\caption{The performance comparison with and without parameter sharing. The results are the average accuracy on 11 datasets of different shots.}
\label{tab:ps}
\end{table}

\begin{table*}[ht]
\centering
\begin{tabular}{lccccccc}
\toprule
\multirow{2}{*}{\textbf{Method}} & \textbf{Source}         & \multicolumn{4}{c}{\textbf{Target}}  & \multirow{2}{*}{\textbf{Average}} & \multirow{2}{*}{\textbf{OOD Average}} \\ \cline{2-6}
& \textbf{ImageNet}& \textbf{-V2}     & \textbf{-S} & \textbf{-A}     & \textbf{-R}     &&    \\ \midrule
CLIP& 66.73   & 60.83   & 46.15& 47.77   & 73.96   & 59.09& 57.18\\ \midrule
CoOp& 71.53   & 64.20   & 47.99& 49.71   & 75.21   & 61.73& 59.28\\
CoCoOp     & 71.02   & 64.07   & 48.75& 50.63   & 76.18   & 62.13& 59.91\\
VPT-Deep   & 70.57   & 63.67   & 47.66& 43.85   & 74.42   & 60.03& 57.40\\ \midrule
MaPLe      & 71.02   & 64.07   & \textbf{49.15}  & \textbf{50.90} & \textbf{76.98} & \underline{62.42}& \textbf{60.28}  \\
UPT & \textbf{72.63} & \underline{64.35}   & 48.66& \underline{50.66}   & 76.24   & \textbf{62.51}& \underline{59.98}\\ \midrule
DCP (ours) & \underline{71.53}   & \textbf{64.50} & \underline{48.77}& 49.40   & \underline{76.50}   & 62.14& 59.79 \\ \bottomrule  
\end{tabular}
\caption{ Domain generalization comparison of DCP with existing approaches. The winners and runners-up are marked in bold font and underlined, respectively.}
\label{tab:xd}
\end{table*}
We intend to learn as few parameters as possible to achieve a transfer of large-scale pre-trained models in downstream tasks. Setting the prompt depth to 9 means that there are 9 CMPA modules, which greatly increases the number of trainable parameters for the model. Hence we conduct the experiment in which the parameters of CMPA are shared across different layers. Table~\ref{tab:ps} shows the average results of different shots on 11 datasets. `PS' is short for `parameter sharing'. It can be observed that on most shots (except for 8 shots) the performance of parameter sharing is higher than non-sharing setting. 

\subsection{Domain Generalization}
After prompt tuning on specific datasets, we do not want to lose the general knowledge of the pre-trained large model.  In this section, we conduct domain adaptation experiments to evaluate the generalization ability of our model DCP. 
\subsubsection{Datasets and Implementation Details}
Following \cite{DBLP:journals/ijcv/ZhouYLL22}, we use ImageNet~\cite{DBLP:conf/cvpr/DengDSLL009} as source domain, and ImageNet V2~\cite{DBLP:conf/icml/RechtRSS19}, ImageNet-Sketch~\cite{DBLP:conf/nips/WangGLX19}, ImageNet-A~\cite{DBLP:conf/cvpr/HendrycksZBSS21} and ImageNet-R~\cite{DBLP:conf/iccv/HendrycksBMKWDD21} as target domains. We train our model on the 16 shots of ImageNet, and test it on other four datasets.
Different from the settings in few-shot task, the training epoch on 16-shot ImageNet in cross domain task is set to 5. We also decrease the prompt length to 8.
\subsubsection{Main Results}
Table~\ref{tab:xd} compares our method DCP with other prompt learning methods on cross-domain tasks.
The compared methods include zero-shot CLIP, unimodal prompt learning methods (CoOp, CoCoOp and VPT-Deep) and multi-modal prompt learning methods (MaPLe and UPT).
The best results on different datasets are in bold, and the second best results are underlined. We can observe that
1) prompt learning does not corrupt the generalization ability of pre-trained large models;
2) multi-modal prompt learning methods outperform unimodal prompt learning methods in generalization performance;
3) our method can get comparable performance as the state-of-the-art methods.

\section{Discussion and Conclusion}
This paper proposes a deeply coupled cross-modal prompt learning method, with a core module cross-modal prompt attention. Our method focuses on optimizing the interaction across different models and layers to address the alignment between vision and language. Experiments on few-shot image classification and domain adaptation evidence that our method can transfer the general knowledge learned by pre-trained foundation models to downstream tasks without penalty of the original generalization ability. Our method provides a strong baseline on few-shot image classification. The deep fusion between visual and language information may enable our approach to have greater potential for complex cross-modal tasks, such as referring expression comprehension~\cite{DBLP:conf/acl/SubramanianMD0022}, image retrieval~\cite{DBLP:conf/cvpr/BaldratiBUB22a} and visual question answering~\cite{DBLP:conf/ijcai/Liu0P022}. We will apply our method to such complicated cross-modal tasks to evaluate its effectiveness in our future work.

\begin{figure}[h]
	\centering
	\includegraphics[width=0.48\textwidth]{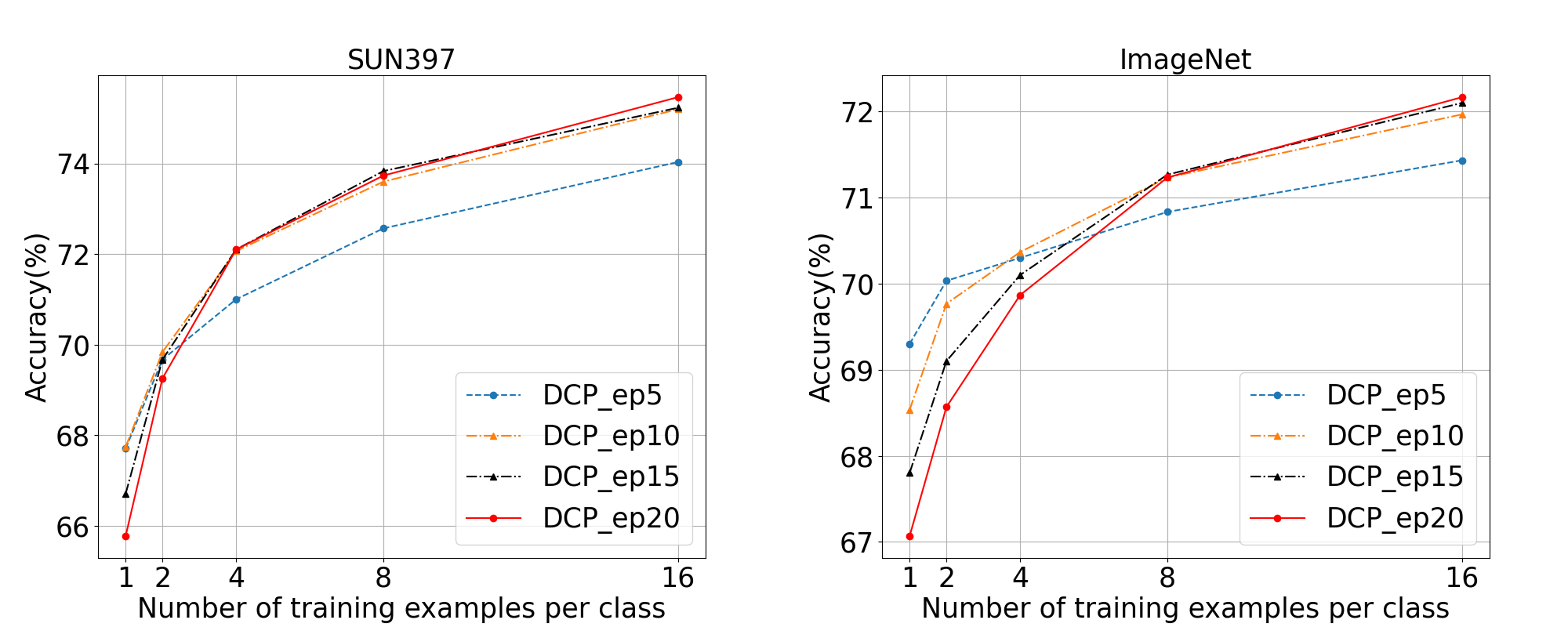}
	\caption{Accuracy comparison of different epochs on Sun397 and ImageNet.}
	\label{fig:limits}
\end{figure}
\section{Limitations}
We discover that for datasets with a relatively large number of categories, our method requires a more delicate setting of epoch under different shots.
Figure~\ref{fig:limits} shows the average results on Sun397 and ImageNet of different epochs. It can be observed that for datasets with a large number of categories (such as Sun397 and ImageNet), as the number of shots decreases, the performance deteriorates with an increase in the number of epochs, which is not evident on the datasets with a small number of categories. We will delve further into this problem to find the reason and solution. 

 \section{Acknowledgement}
 We would like to thank anonymous reviewers for their insightful comments to help improve the paper. This publication has emanated from research conducted with the support of SenseTime Research and Hetao Shenzhen-Hong Kong Science and Technology Innovation Cooperation Zone (HZQB-KCZYZ-2021045.

\bibliography{anthology,custom}
\bibliographystyle{acl_natbib}

\appendix



\end{document}